\documentclass[final,3p,times]{elsarticle}
\makeatletter
\def\ps@pprintTitle{%
 \def\@oddfoot{\footnotesize\itshape%
   \begin{minipage}[t]{\textwidth}
   \centering Preprint \hfill \@date 
   \end{minipage}}%
 \let\@evenfoot\@oddfoot}
\makeatother

\usepackage{graphicx}
\usepackage{color,soul} 
\usepackage{array}
\usepackage{amsmath}
\usepackage{amsfonts}
\usepackage{amssymb}
\usepackage{amsthm}
\usepackage{psfrag}
\usepackage{epstopdf}
\usepackage{subcaption}
\usepackage{caption}
\usepackage{booktabs}
\usepackage{hyperref}
\usepackage{latexsym}
\usepackage{kantlipsum}
\usepackage{balance}
\usepackage{lscape}
\usepackage{rotating}
\usepackage{multicol}
\usepackage{multirow,bigdelim}
\usepackage{color, colortbl}
\usepackage{threeparttable}
\usepackage{float}
\usepackage{algorithm, algorithmicx, algpseudocode}
\usepackage{footnote}

\usepackage{xcolor}

\definecolor{lightblue}{HTML}{ADD8E6}

\hypersetup{
    colorlinks=true,
    linkcolor=lightblue,
    urlcolor=blue,
    citecolor=lightblue
}

\newcommand{\eval}[2][\right]{\relax\ifx#1\right\relax \left.\fi#2#1\rvert}

\newcommand{\bu}{\boldsymbol{u}}

\newcommand{\btx}{\boldsymbol{\Tilde{x}}}

\newcommand{\bT}{\boldsymbol{T}}

\newcommand{\bX}{\boldsymbol{X}}

\makeatletter
\newcommand{\Rmnum}[1]{\expandafter\@slowromancap\romannumeral #1@}
\makeatother

\newcommand{\bg}{\boldsymbol{g}}

\newcommand{\bK}{\boldsymbol{K}}

\newcommand{\uhat}{\hat{\bu}}

\newcommand{\bq}{\boldsymbol{q}}

\begin{document}

\begin{frontmatter}

% --------------- Title ----------------------------
\title{\textcolor{black}{A novel data generation scheme for surrogate modelling with deep operator networks}}
\tnotetext[t1]{}
% --------------------------------------------------
%\title{A novel stress-based formulation from mixed Hellinger-Reissner principle in three-dimensional isogeometric analysis (Draft Title)}

%% use optional labels to link authors explicitly to addresses:
%\author[label1]{Dhiraj S. Bombarde}  
% ---------------- author 1 ----------------------
\author[label1]{Shivam Choubey}
%\ead{shivamchb42@gmail.com}
% --------------- author 2 -----------------------
\author[label2]{Birupaksha Pal} 
%\ead{Birupaksha.Pal@in.bosch.com}
% --------------- author 3 -----------------------
\author[label1]{Manish Agrawal\corref{cor1}}
\ead{manish.agrawal@iitrpr.ac.in}
% ------------------------------------------------ 
\cortext[cor1]{Corresponding author}

\address[label1]{Department of Mechanical Engineering, Indian Institute of Technology Ropar, Rupnagar-140001, Punjab, India}
\address[label2]{Robert Bosch Research and Technology Center India, Bengaluru 560095, Karnataka, India}

\begin{abstract}

Operator-based neural network architectures such as DeepONets have emerged as a promising tool for the surrogate modeling of physical systems. In general, towards operator surrogate modeling, the training data is generated by solving the PDEs using techniques such as Finite Element Method (FEM). The computationally intensive nature of data generation is one of the biggest bottleneck in deploying these surrogate models for practical applications. In this study, we propose a novel methodology to alleviate the computational burden associated with training data generation for DeepONets. Unlike existing literature, the proposed framework for data generation does not use any partial differential equation integration strategy, thereby significantly reducing the computational cost associated with generating training dataset for DeepONet. In the proposed strategy, first, the output field is generated randomly, satisfying the boundary conditions using Gaussian Process Regression (GPR). From the output field, the input source field can be calculated easily using finite difference techniques.  The proposed methodology can be extended to other operator learning methods, making the approach widely applicable. To validate the proposed approach, we employ the heat equations as the model problem and develop the surrogate model for numerous boundary value problems.
\end{abstract}


%Computational modeling, enabled by techniques like finite element and finite volume methods, serves as a computational alternative for simulating physical systems governed by partial differential equations (PDEs). However, the intensive computational demands of these techniques limit their direct application in scenarios like parametric optimization and real-time prediction. Machine learning-based surrogate models based on operator learning offer a viable solution. 
\end{frontmatter}

\section{Introduction}\label{Introduction}
Techniques such as finite element and finite volume methods are extensively used for the simulation of physical systems by solving the underlying partial differential equations (PDE). However, due to intensive computational requirements, it is not feasible to deploy these techniques directly in numerous cases, such as parametric optimization, real-time prediction for control applications, etc. Machine learning-based surrogate models offer an alternate way for simulation of the physical systems in an efficient manner. Deep learning, due to its ability to model any arbitrary input-output relationship in an efficient manner is the most accepted choice for surrogate modelling.  In general, these surrogate models are data driven models, where the simulation/experimental data is used for the training purpose.  Once the surrogate model is trained, it can be  used to predict the system output for unobserved data with minimal computational effort.

For  surrogate modelling, both vanilla and specialized neural networks such as convolution neural networks have gained immense popularity in both scientific  as well as for industrial applications \citep{Khorrami2023, article}. Further, recently in \citep{kovachki2023neural}, operator learning, a new paradigm in deep learning is proposed. In literature, various operator learning techniques are proposed, like deep operator networks (DeepONets)\cite{lu2020deeponet}, Laplace Neural operators (LNO)\cite{cao2023lno}, Fourier Neural operators (FNO)\cite{li2021fourier} and General Neural Operator Transformer for Operator learning (GNOT)\cite{hao2023gnot}. In this paper, we focus on DeepONets as an operator learning technique and show a novel way on how to reduce the computational cost associated with training the model. DeepONet is based on the lesser known cousin of the ‘Universal Approximation Theorem (UAT)’ called ‘Universal Approximation theorem for operator (UATO)’\cite{chen1995universal} which states that a neural network with a single hidden layer can accurately approximate any nonlinear continuous functional or operator. This is a powerful result which in essence says that as neural networks can approximate any function in a similar vein operator networks can be used to approximate any functional too. It is a purely data driven approach which inherently learns about the operator transforming input function space to output function space rather than learning about the relation between the input and output field. This allows DeepONet to achieve a far better generalization and makes it a far better choice for data-driven surrogate modelling \citep{shukla2023deep}.

%that offers advantage such as generalizability   over the traditional neural network architectures. Operator learning aims to predict the solution of the partial differential equation by learning the underlying differential operator.  
Operator learning is especially usefully for surrogate modelling where the output field  as a function of a source term is solicited. For example, in \citep{koric2023data}, for the heat conduction problem, authors have predicted the temperature field as a function of the heat source with the help of DeepONet network. 
%The Physics Informed neural networks (PINNs) offers a alternative viw for the surrare a new paradigm of surrogate modelling methods which aims at addressing the issue of interpretability and data requirement of traditional neural nets. PINNs integrates the knowledge of governing PDEs of physical systems by modifying the cost function to include the information from the equation resulting in little or no measurement data requirement for training. However, PINNs have a limited flexibility in terms of model generalization, i.e, in case the inputs or source terms, parameters or the domain changes the trained model might not be able to make reliable predictions for the altered system. Generally, the problem can be addressed by retraining the model for the updated system settings\cite{koric2023data}. But for most industrial applications this is a bottleneck as most models are integrated into complex pipelines and collecting data again for retraining can be significantly expensive which could even render the whole modelling exercise futile from the cost perspective.\\
DeepONet based surrogate model are also shown to be robustly applied for applications like gravity pendulum with an external force, diffusion-reaction system with very small generalization error~\cite{lu2020deeponet}.  The challenging problem of fracture mechanics  has been addressed in \citep{goswami2022physics} which uses DeepONet as a surrogate model for predicting crack path in quasi-brittle materials. Domain based transfer learning of DeepONet has also been explored in \cite{goswami2022deep} for Darcy flow problem, elasticity model and brusselator diffusion-reaction system. The use of DeepONet based surrogate model for reliability analysis by computing First Passage Failure Time (FPFT) for SDOF Bouc Wen System, 5-DOF Non-linear System, etc. has also been explored\cite{garg2022assessment}. Integration of many machine learning strategies like POD (proper orthogonal decomposition)\cite{liu2022causality}, CNN (convolutional neural network) \cite{goswami2022deep, Oommen2022}, Bayesian training strategy \cite{lin2021accelerated}, etc have also been investigated. PINNs can also be combined with DeepONets for a new architecture called Physics Informed DeepONets\cite{goswami2022physicsinformed, wang2021learning} where in the PINN architecture,  DeepONet is used instead of ANN , with most of the remaining algorithm unchanged. The DeepONet in the PINN model architecture adds to the generalizability of the model.

The above citations show the flexibility and generalizability capability of DeepONet for various use cases. However, it is crucial to emphasize that any operator learning needs to be trained for various input scenarios, turning them into data-intensive processes. Though DeepONets can be used as a data based modelling algorithm for situations where governing equations are not available, one of the main application of them is to use as a surrogate model for conventional PDE solvers such as FEM. To build a good surrogate model for a complex physical system, a fairly good amount of training data has to be generated from simulations. Creating an adequate dataset for training surrogate models is computationally demanding and often acts as a primary bottleneck. Hence, there is a need for some novel modifications that can help alleviate the computational costs associated with training data.

In this manuscript, we propose a novel methodology for data generation for surrogate modelling for DeepONets. The proposed approach is motivated by the insight that for any PDE, given the primary field variable, it is easier to calculate the source field compared to finding the  primary field from a given source field. For example, in the case of a steady-state conduction problem governed by the heat equation, it is straightforward to calculate the heat source from a given temperature field, whereas finding the temperature field from a  given heat source is much more intricate. Even in the case of numeric, the computation of the derivative using technique such as  central difference is relatively straightforward as compared to numerically integrating the partial differential equations using techniques such as FEM. Thus, towards the generation of the training data, we proposed the generation of the primary field first, thereby calculating the heat source from it. The main challenge of the generation of the primary field directly is the requirement to satisfy the boundary conditions. We propose using the Gaussian process regression (GPR) technique to satisfy the boundary conditions of the primary field.  GPR \citep{Wang_2023, rasmussen2006gaussian} is a powerful and flexible non-parametric Bayesian machine learning technique widely used for regression and uncertainty estimation. In this manuscript, we have focused only on the Dirichlet boundary conditions. Once the primary field satisfying the boundary condition is obtained, finite difference techniques can easily calculate the heat source. 

Since both the GPR as well as finite difference scheme are computationally less intensive  compared to the FEM, the proposed methodology can provide an efficient way of  the generation of training data for operator-based surrogate modelling. This method can be utilized not only for DeepONets but also can be extended for other operator learning methods mentioned earlier. In this work, we have used the heat equations as the model problem to verify the proposed approach. Numerous numerical problems have been solved in the manuscript to demonstrate the efficacy of the proposed approach. We have solved the problem for homogeneous as well as heterogeneous boundary problems. The overall performance in various numerical examples shows that the proposed framework is general and can be applied to operator learning techniques.

The paper is divided into four sections, where \hyperref[methodology]{Section 2} provides an overview of the DeepONet architecture and details the process of preparing training and testing data for DeepONet. \hyperref[proposition]{Section 3} describes the proposed framework for the data generation using GPR and finite difference method. In \hyperref[Results]{section 4}, we demonstrate the performance of the proposed approach on various numerical examples.

\section{Deep operator network (DeepONet)}\label{methodology}
In this section, we present the preliminary of the DeepONet based surrogate modelling, for more details readers can refer \citep{lu2020deeponet}. Let $\boldsymbol{U}$ be the input function space of the system and $\boldsymbol{S}$ be the solution space, $G$ be an differential operator transforming $\boldsymbol{U}$ to $\boldsymbol{S}$, $u$ be a function belonging to $\boldsymbol{U}$, and $\btx$ be a point in the domain of $G(u)$. Thus, it can be represented mathematically as:
\begin{equation}
    G: U \rightarrow S,\quad S \equiv G(u)(\btx)
    \label{e1}
\end{equation}
The DeepONet based operator learning learn the mapping between $U$ and $S$ in an efficient manner. For example, let us consider the case of  2-D steady state heat conduction, where the the heat source \(q\)  and the temperature field \(T\) are related by the given  PDE:
\begin{equation}
\begin{aligned}
    \frac{\partial}{\partial x}\left(k \frac{\partial T}{\partial x}\right) + \frac{\partial}{\partial y}\left(k \frac{\partial T}{\partial y}\right) + q(x,y) &= 0, \quad \text{on} \; \Omega \\
    T&=\bar{T}, \quad \text{on} \; \partial\Omega
 \end{aligned}
    \label{e_2_2}
\end{equation}
where: $k =$ thermal conductivity and $\btx=(x,y)$ are spatial coordinates.\\
Thus in case of the 2-D steady state heat conduction equation, the heat source \(q\) is the input field \(u\), and the temperature field \(T\) is the output field \(G(u)(\btx)\). The task of the DeepONet is to learn the integration operator of the above partial differential equations and predict the temperature field for a given heat source.

\subsection{Architecture}
As discussed above, let us consider that output field \( G(u)(\btx)\) is a function of a source \(u\) and the spatial coordinates \(\btx=(x,y)\). For prediction of the output field \( G(u)(\btx)\), the DeepONet architecture intrinsically uses two independent vector fields  \(\boldsymbol{f}\) and  \(\bg\) of identical dimension \(c\). Wherein, \(\boldsymbol{f}(u)\) is only dependent on the source field \(u\), while the \(\bg(x,y)\) is only dependent on the spatial coordinates. With these function choices, the output from the DeepONet network  can be  written in the following functional form:
  \begin{equation}
    G(u)(\btx) = \sum_{i=1}^{c}f_i(u)\cdot g_i(\btx) 
    \label{e2}
\end{equation}
As can be seen from the above expression the output from the DeepONet network is given as the dot product of the two vectors fields \(\boldsymbol{f}\) and \(\bg\). In \citep{lu2020deeponet}, the universal approximation theorem for operators is presented that shows that the above functional form~\autoref{e2} can represent any continuous functional operator. This novel architecture of the DeepONet allows the split learning among the two functions and ultimately enable it to efficiently learn the underlying operator. The vector field \(\boldsymbol{f}\) focuses on capturing the dependence of the output field on the source term \(u\), while \(\bg\) captures the spatial dependency of the output field. 

\begin{figure}[!ht]
    \centering
    \includegraphics[width=0.7\linewidth]{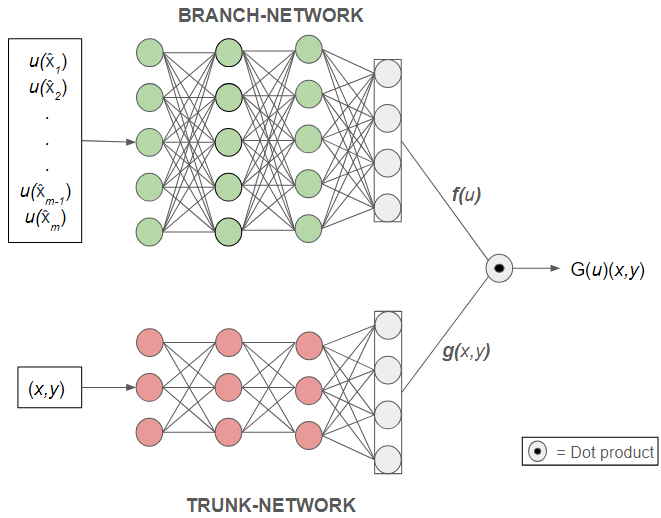}
    \caption{DeepONet Architecture consists of two sub neural networks namely branch and trunk network.}
    \label{Deeponet}
\end{figure}

The functional form given in equation~\autoref{e2} is captured in neural network architecture by having two sub neural networks called branch and trunk networks \citep{lu2020deeponet}. A typical DeepONet architecture is shown in the~\autoref{Deeponet}  The input to the branch network is the source field \(u\) at sampled points and the output from this network is a  vector \(\boldsymbol{f}\). The second network which is called the trunk network, takes coordinates (\(x,y\)) as input, while the output is vector field \(\bg\). The output of the both of the sub networks  is of the size c.  The dot product of the \(\boldsymbol{f}\) and \(\bq\) gives an approximation  of the quantity $G(u)(\btx)$ or the output of the system. 

%Thus the branch network focuses on leaning the dependence of the output field on the input field \(u\), while the trunk network captures the dependency of the spatial dependency of the output field. This novel architecture of the deep-onet allows the split learning among the two networks. This prior of split learning ultimately enable to efficeintly learn the underlying operator.

%Deeponet architecture consists of the two neural networks namely; \textbf{branch-net} and \textbf{trunk-net}. Both, the branch and trunk network, provides the vector of the identical length as output.  The final output from the deeponet, is given by the dot product of the branch and trunk network output vectors. 

  %This network in essence learns the effect of the input function 'worked upon' by the operator at the evaluation points. As an output both the networks return a 'p' numbered embeddings of their input features. 
 %The output of Deeponet can be mathematically represented as:

%where $b_i$ is branch-net output, $t_i$ = trunk-net output and c= number of common units.\\
% Or in other terms, simplifying \autoref{e1_1} for Deeponet, (\( \sum_{i=1}^{n} c_k^i \sigma\left(\sum_{j=1}^{m} \xi_k^{ij}u(x_j) + \theta_k^i\right)\)) refers to output of \textbf{branch-net} of Deeponet and (\( \sigma(w_k \cdot y + \zeta_k)\)) refers to output of \textbf{trunk-net} and their dot product can successfully approximate any non-linear operator, G(u)(y).

\subsection{Preparation of training and testing data}
The convergence of DeepONet is proven in~\cite{lu2020deeponet} with Gaussian Random Fields (GRF) as the  choice of input function space. Thus, in general for training purposes, the input field \(u\) is generated as GRF. GRF~\citep{liu2019advances} are a special type of random fields characterised by Gaussian probability distribution. Any finite collection of points in GRF follows a multivariate Gaussian distribution. The multivariate Gaussian distribution are characterised by the kernel function ($\bK$). This kernel function defines the covariance structure of the GRF, capturing the spatial dependence between different points in the field. Specifically, it determines how the values of the field at different locations are correlated. 

For any field ($u$)  belonging to GRF, the input field is measured at \(m\)  number of finite discretized  points $\hat{x}_i=(x_i,y_i),i=1,2...m$. These input sampling points are called the sensor points. The discretized from (\(\hat{\bu} \)) can be given by the following multivariate Gaussian distribution:
\begin{equation}
    \hat{\bu} \sim \mathcal{N}(\mu, \bK)\quad 
    \label{e3}
\end{equation}

For training purpose, consider that for \(n\) independent input fields \((\uhat^{(1)},\uhat^{(2)},...,\uhat^{(n)}\)) are generated. Further after generation of the input field, output field labels are required to be computed. To obtain the output field, typically techniques such as FEM are deployed to integrate the underlying partial differential equation. Lets consider that the corresponding output fields \(G(\uhat^{(1)})(x,y),G(\uhat^{(2)})(x,y),...,G(\uhat^{(n)})(x,y)\) are generated through any numerical integration scheme.    We consider that for $i^{th}$ output function \(G(\uhat^{(i)})(x,y)\), the output is evaluated at \(p\) number of points \((\hat{y}_j^{(i)}=(x_j,y_j),j=1,2...p.)\). These \(p\)locations are called the output locations. With these considerations, the training dataset of DeepONet consists of the following matrices:
\begin{itemize}
    \item \textbf{Input function matrix:} It is $(np\times{m})$ matrix consisting of $n$ different input functions ($\hat{\bu}^{(i)}$), discretized at \(m\) number of points. For every input function, the output field values needs to be provided at \(p\) output locations.
    \item \textbf{Output location matrix:} It is $(np\times1)$ matrix consisting of the location where the output is required for each of the \(n\) functions.
    \item \textbf{Output function matrix:} It is $(np\times1)$ matrix consisting of the value of output for every output location and input function combination.
\end{itemize}
\begin{equation}
    \left(\begin{array}{cccc}
    \hat{u}^{(1)}(\hat{x}_1) & \hat{u}^{(1)}(\hat{x}_2) & \ldots & \hat{u}^{(1)}(\hat{x}_m) \\
   \vdots & \vdots & \vdots & \vdots \\
      
    \hat{u}^{(1)}(\hat{x}_1) & \hat{u}^{(1)}(\hat{x}_2) & \ldots & \hat{u}^{(1)}(\hat{x}_m) \\
 \vdots & \vdots & \vdots & \vdots \\
    \hat{u}^{(1)}(\hat{x}_1) & \hat{u}^{(1)}(\hat{x}_2) & \ldots & \hat{u}^{(1)}(\hat{x}_m) \\
    \cdot & \cdot & \dots & \cdot \\
      \cdot & \cdot & \dots & \cdot \\
        \cdot & \cdot & \dots & \cdot \\
     \hat{u}^{(i)}(\hat{x}_1) & \hat{u}^{(i)}(\hat{x}_2) & \ldots & \hat{u}^{(i)}(\hat{x}_m) \\
  \cdot & \cdot & \dots & \cdot \\
      \cdot & \cdot & \dots & \cdot \\
        \cdot & \cdot & \dots & \cdot \\
     \hat{u}^{(n)}(\hat{x}_1) & \hat{u}^{(n)}(\hat{x}_2) & \ldots & \hat{u}^{(n)}(\hat{x}_m) \\
   \vdots & \vdots & \vdots & \vdots \\
    \hat{u}^{(n)}(\hat{x}_1) & \hat{u}^{(n)}(\hat{x}_2) & \ldots & \hat{u}^{(n)}(\hat{x}_m) \\
 \vdots & \vdots & \vdots & \vdots \\
    \hat{u}^{(n)}(\hat{x}_1) & \hat{u}^{(n)}(\hat{x}_2) & \ldots & \hat{u}^{(n)}(\hat{x}_m) \\
\end{array}\right)\quad , \quad
\left(\begin{array}{c}
    \hat{y}^{(1)}_1 \\
    \vdots  \\
    \hat{y}^{(1)}_j \\
    \vdots  \\
    \hat{y}^{(1)}_p \\
    \cdot  \\
    \cdot  \\
    \cdot  \\
    \hat{y}^{(i)}_j  \\
    \cdot  \\
    \cdot  \\
    \cdot  \\
    \hat{y}^{(n)}_1 \\
    \vdots  \\
    \hat{y}^{(n)}_j \\
    \vdots  \\
    \hat{y}^{(n)}_p \\
\end{array}\right)\quad , \quad
\left(\begin{array}{c}
    G(\uhat^{(1)})(\hat{y}^{(1)}_1)  \\
    \vdots  \\
    G(\uhat^{(1)})(\hat{y}^{(1)}_j)  \\
    \vdots \\
    G(\uhat^{(1)})(\hat{y}^{(1)}_p)  \\
    \cdot  \\
    \cdot  \\
    \cdot  \\
    G(\uhat^{(i)})(\hat{y}^{(i)}_j)  \\
    \cdot  \\
    \cdot  \\
    \cdot  \\
    G(\uhat^{(n)})(\hat{y}^{(n)}_1)  \\
    \vdots  \\
    G(\uhat^{(n)})(\hat{y}^{(n)}_j)  \\
    \vdots \\
    G(\uhat^{(n)})(\hat{y}^{(n)}_p)  \\
\end{array}\right)
\label{datamat}
\end{equation}
In the above representation, $\uhat^{(i)}$ refers to the $i^{th}$ input function, $y^{(i)}_j$ refers to the $j^{th}$ output location for $i^{th}$ input function and $G(u^{(i)}_j)(y^{(i)}_j)$ refers to the output corresponding to $j^{th}$ output location for $i^{th}$ input function.
The architecture allows flexibility in choosing output location meaning the training and testing output location can be different. This property of DeepONet makes its prediction mesh-independent.

%[u(x_1),\ldots, u(x_m)]

\section{Proposed data generation framework:}\label{proposition}

As described in the previous section, the DeepONet network can learn the underlying partial differential equation integrator operator from the underlying training data. The training data for operator learning consists of the source field \(u\) and primary variable pairs \autoref{datamat}. In existing work, first the random source fields are generated and corresponding primary fields are obtained by integrating the underlying PDE.  Here, in the proposed framework, we propose an alternative framework, wherein, first the random primary field is generated and corresponding source field is calculated with the help of the partial derivatives of the primary field.  

\begin{figure}[ht]
\centering
\begin{subfigure}[b]{0.49\textwidth}
  \centering
  \includegraphics[width=0.92\textwidth]{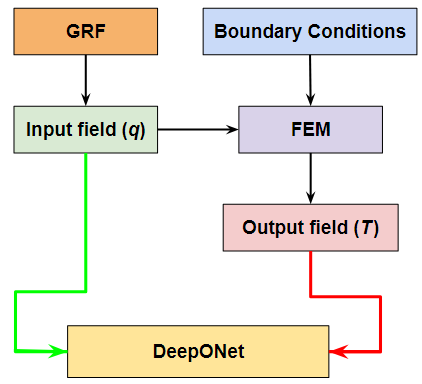}	% here, 'width' is used to define size of the figure
  \caption{\label{5_1a}}
\end{subfigure}
\begin{subfigure}[b]{0.49\textwidth}
  \centering
  \includegraphics[width=0.99\textwidth]{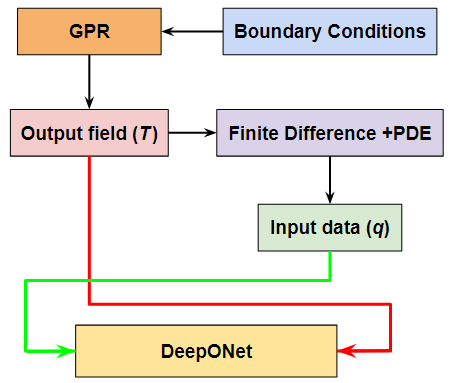}
  \caption{\label{5_1b}}
\end{subfigure}
\caption{Flowchart of data generation for steady state heat conduction problem : (\subref{5_1a}) existing approach in literature, (\subref{5_1b}) proposed framework. Unlike existing work the proposed framework does not require the use of FEM technique. }
\label{5_1}
\end{figure}

The proposed framework is inspired from the simple understanding that it is easier to calculate the derivatives compared to performing the integration. This is especially true for PDE, where computation of the partial derivatives is much more straightforward compared to integrating them.   For example, in case of steady state conduction problem, governed by \autoref{e_2_2}, it is straightforward to calculate the heat source \(q\) from a given temperature field \(T\) that is consistent with the boundary conditions, whereas to find the temperature field from a  given heat source \(q\) is much more difficult. Even in numerical implementation of  discretized form, the computation of the derivative using technique such as  central difference is rather straightforward as compared to numerically integrating the partial differential equations using techniques such as FEM. 
The overall proposed framework flow chart and its comparison with the existing literature is shown in~\autoref{5_1}. As can be seen that using the proposed framework, pairs of input(heat flux)-output(temperature)  are obtained without the need of technique such as FEM. These input-output pairs can be used for the training and testing of operator networks. 
 
\begin{figure}[!ht]
     \centering
     \includegraphics[trim= {1.2cm 0.6cm 1.6cm 1.4cm},clip, width=0.5\textwidth]{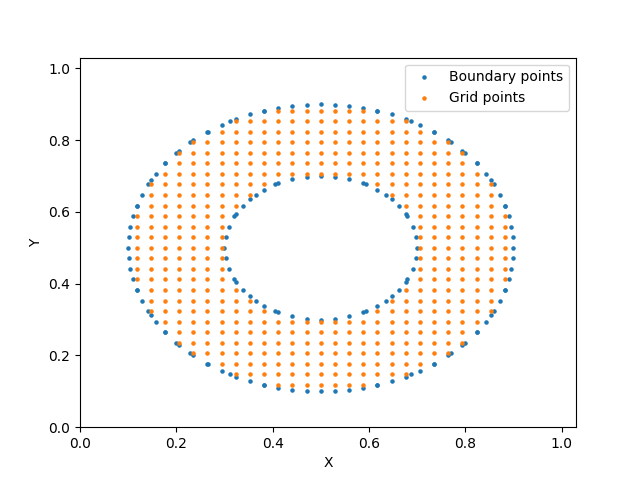}
     \caption{Example for grid points in a domain: Grid points for an annular domain are shown. The grid points can be used as sensor as well as output points. }
     \label{5_0_1}
 \end{figure}

To explain the proposed framework, we take help of the example of steady state heat equation \autoref{e_2_2}. In the proposed framework for the data generation, we first discretize the domain in a grid. For example, a typical grid  for a annual geometry is shown in the~\autoref{5_0_1}. Let $\hat{\bX}$ consists of all the coordinates of the grid points ($\hat{X}_i=(x_i,y_i)$). Further, let us consider that $\hat{\bT}$ represent the discretize temperature field, and contain the temperature values $\hat{T}_i=T(x_i,y_i)$ at the grid points. Similarly,  $\hat{\bq}$ represent the discretize heat source, and contain the heat source values $\hat{q}_i=q(x_i,y_i)$ at the grid points. The grid points can be used as output locations as well as sensor location. Thus we strive to find the values of the discretize random temperature $\hat{\bT}$ and heat source $\hat{\bq}$.  

\subsection{Generation of random temperature field}\label{gpr}
In the proposed framework, as stated above, we first generate the random random temperature field, and the heat source is computed form the obtained field. For generation of the random temperature field,  the main challenge is in ensuring the given boundary conditions. In the manuscript our focus is only on the Dirichlet boundary conditions. To ensure that the generated random temperature field satisfy the Dirichlet boundary conditions, we leverage the capabilities of Gaussian Process Regression (GPR). GPR uses the Bayesian framework to craft Gaussian Random Fields (GRFs) in adhering the value of random field at the specified points. Here, we use the GPR to force the temperature field to meet the boundary conditions. As per standard GPR process, we first assume a prior distribution of the temperature field. For training a DeepONet network, we need the discretized random temperature field ($\hat{\bT}(\hat{\bX})$) i.e. the temperature values at the grid points. The prior of the discretized temperature field    $\hat{\bT}(\hat{\bX})$ is generated from the multivariate Gaussian distribution as follow: 
  \begin{equation}
        \hat{\bT}(\hat{\bX}) \sim \mathcal{N}(\mu(\hat{\bX}), \bK),
  \end{equation}
  
    where:
      \begin{itemize}
        \item \(\mu(X)\) is the mean of the prior field. In general, we can take \(\mu(X)\)=0. 
        \item \(\bK\) is the covariance matrix, where \(K_{ij} = k(\hat{X}_i, \hat{X}_j)\) is determined by kernel function.
    \end{itemize}
The kernel function, $k(\hat{X}i, \hat{X}_j)$, encodes the spatial correlation between different input points. In this work, we have  taken the Radial Basis Function (RBF) kernel also known as the Gaussian kernel as our choice. The RBF kernel is defined as:
\begin{equation}
    K_{ij} = k(\hat{X}_i,\hat{X}_j) = \sigma^2\exp\left(-\frac{\|\hat{X}_i - \hat{X}_j\|^2}{2l^2}\right)
    \label{rbf}
\end{equation}
where, \(l\) is a characteristic length scale parameter which determines the smoothness of the distribution and $\sigma$ is the spread parameter which determines the magnitude of the field.

The above generated temperature prior does not satisfy the boundary condition. However, the underlying temperature field has to satisfy the boundary conditions in order to be physically consistent with the specified problem statement. In this work as stated above, we have consider only the Dirichlet boundary conditions and thus the temperature is specified at all the boundaries~\autoref{e_2_2}. Let the boundary of the domain is discretized with  \(n_o\)  number of points, and  $\hat{\bX_b}$ contain the coordinates of the these  boundaries points.  Based on these specified temperature values $\bar{\bT}(\hat{\bX_b})$ at the boundary points $\hat{\bX_b}$ the posterior distribution is a normal distribution  \(\bT(\hat{\bX}) | \bar{\bT}(\hat{\bX_b}) \sim \mathcal{N}(\mu^*, \bK^*)\), with mean and variance given by the following expressions:
\begin{equation}
    \begin{aligned}
        \mu^*(\hat{\bX}) &= \mu(\hat{\bX}) + \bK(\hat{\bX}, \hat{\bX_b}) (\bK(\hat{\bX_b},\hat{\bX_b}) + \sigma_{\epsilon}^2I)^{-1} (\bar{\bT}(\hat{\bX_b}) - \mu(\hat{\bX_b}))\\
        \bK^*(\hat{\bX},\hat{\bX}) &= \bK(\hat{\bX}, \hat{\bX}) - \bK(\hat{\bX}, \hat{\bX}_b) (\bK(\hat{\bX}_b,\hat{\bX}_b) + \sigma_{\epsilon}^2I)^{-1} \bK(\hat{\bX}_b, \hat{\bX})\
    \end{aligned}
    \label{eq_gpr}
\end{equation}
where,
\begin{itemize}
    \item \(\bK(\hat{\bX}, \hat{\bX}_b)\) is the covariance matrix between grid values \(\hat{\bX}\) and the boundary data \(\hat{\bX_b}\).
     \item  \(\bK(\hat{\bX_b}, \hat{\bX})\) is the covariance matrix between the boundary data and the grid values \(\hat{\bX}\) .
    \item \(\bK(\hat{\bX_b}, \hat{\bX}_b)\)  is the covariance matrix at the boundary points. The size of this matrix is \(n_o\times n_o\).
    \item \(\bK(\hat{\bX}, \hat{\bX})\)  is the covariance matrix at the grid points. The size of this matrix is \(n_p\times n_p\).
    \item \(\sigma_{\epsilon}^2\) is the noise taken.  \(I\) is the identity matrix.
\end{itemize}

%Where:
%\begin{itemize}
   % \item \(X\) is the vector consisting of the boundary observed points.
   % \item \(x^*\) refers to output location, which is any point in the domain where the temperature field is discretized
   % \item \(\sigma^*\) is the standard deviation (or variance) of the posterior distribution at the  point \(x^*\).
      
\begin{figure}[ht]
\centering
\begin{subfigure}[b]{0.49\textwidth}
  \centering
  \includegraphics[trim= {2.05cm 1.5cm 0cm 0cm},clip,width=0.45\textwidth]{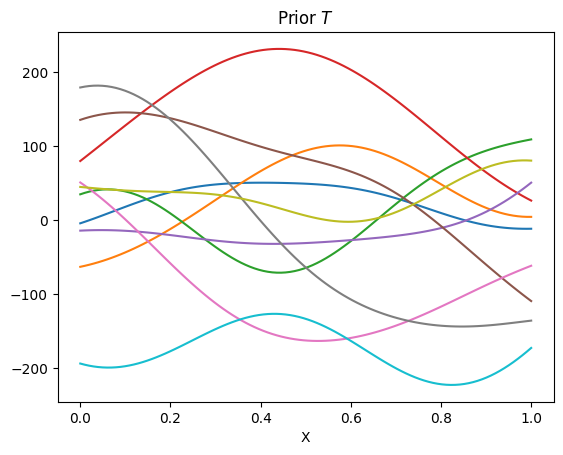}
  \includegraphics[trim= {2.05cm 1.5cm 0cm 0cm},clip,width=0.45\textwidth]{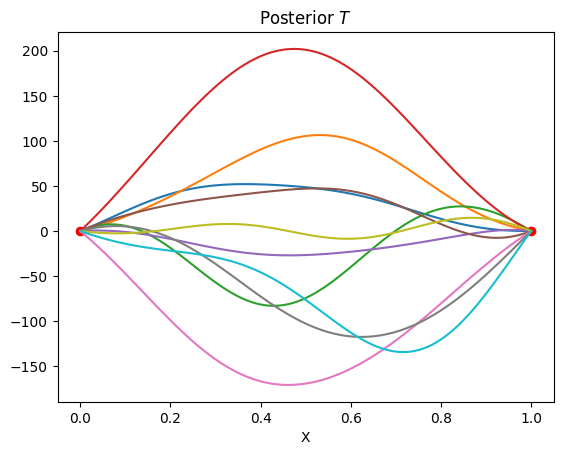}
  \caption{\label{5_0a}}
\end{subfigure}
\begin{subfigure}[b]{0.49\textwidth}
  \centering
  \includegraphics[trim= {2.7cm .6cm 12cm 0cm},clip,width=0.45\textwidth]{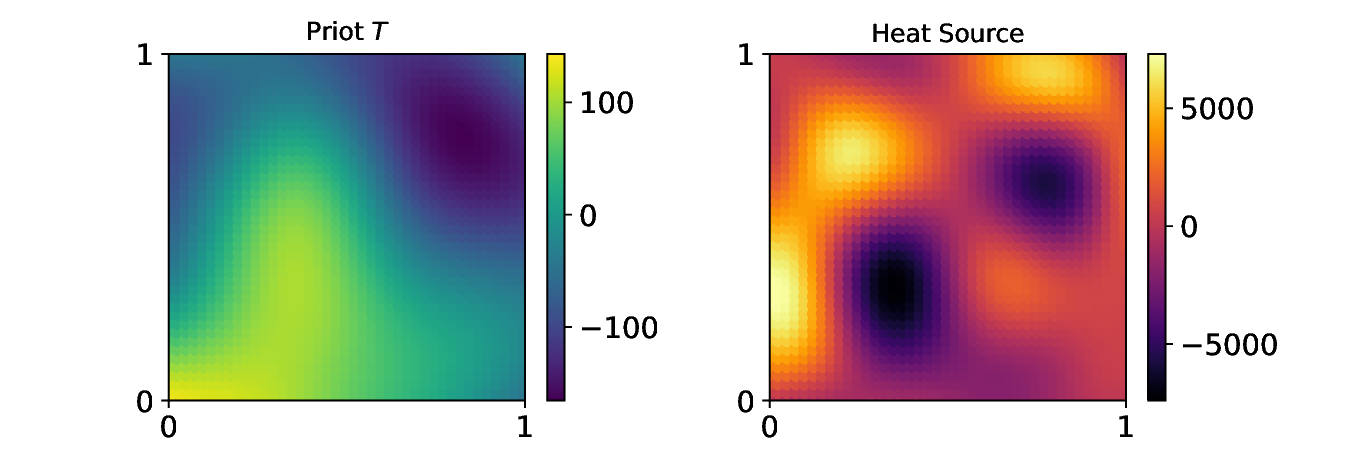}
  \includegraphics[trim= {2.7cm .6cm 12cm 0cm},clip,width=0.45\textwidth]{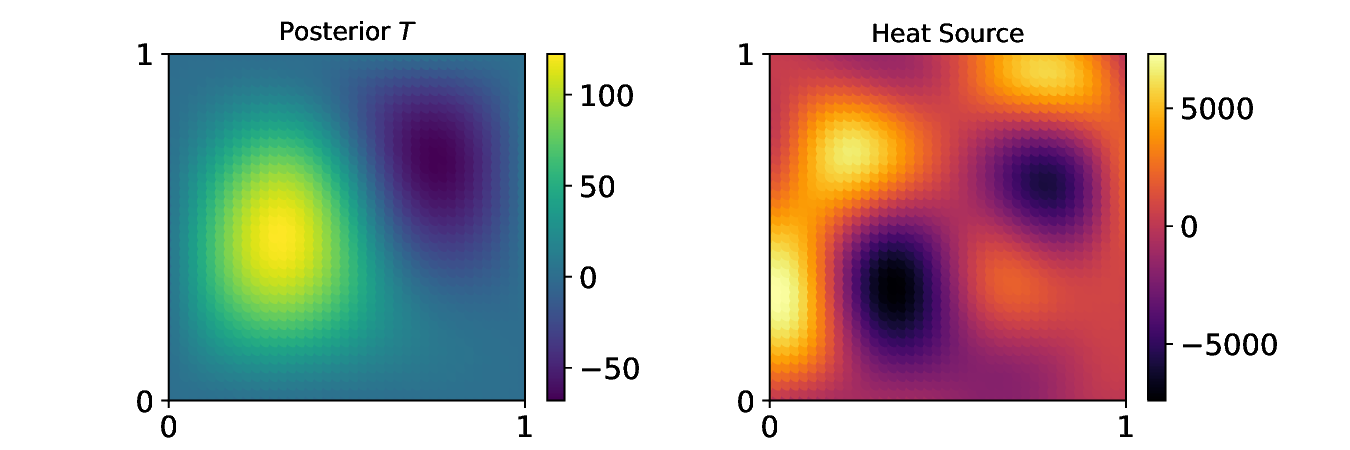}
  \caption{\label{5_0b}}
\end{subfigure}
 \caption{ A sample representation prior and posterior temperature field for a) 1-D b) 2-D heat conduction. Figure shows that temperature generated from posterior distribution using GPR is able to satisfy the boundary conditions.}
\label{5_0}
\end{figure}

\begin{figure}[ht!]
    \centering
    \begin{subfigure}[b]{0.33\textwidth}
        \centering
        \includegraphics[trim= {2cm 0 12cm .75cm},clip, width=1\textwidth]{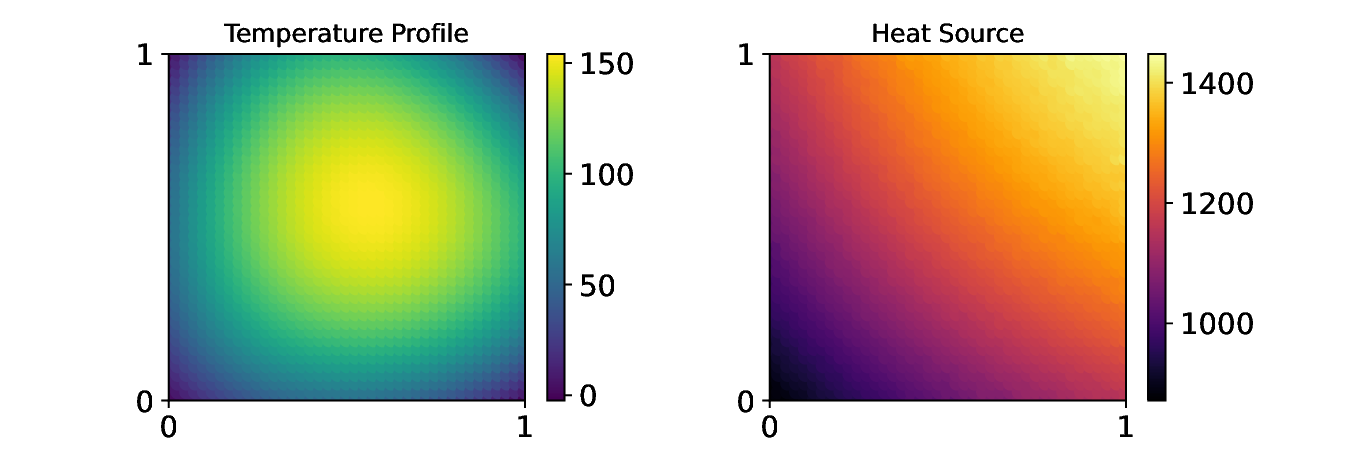}
        \caption{\label{6_2ta}}
    \end{subfigure}
    \begin{subfigure}[b]{0.33\textwidth}
        \centering
        \includegraphics[trim= {2cm 0 12cm .75cm},clip, width=1\textwidth]{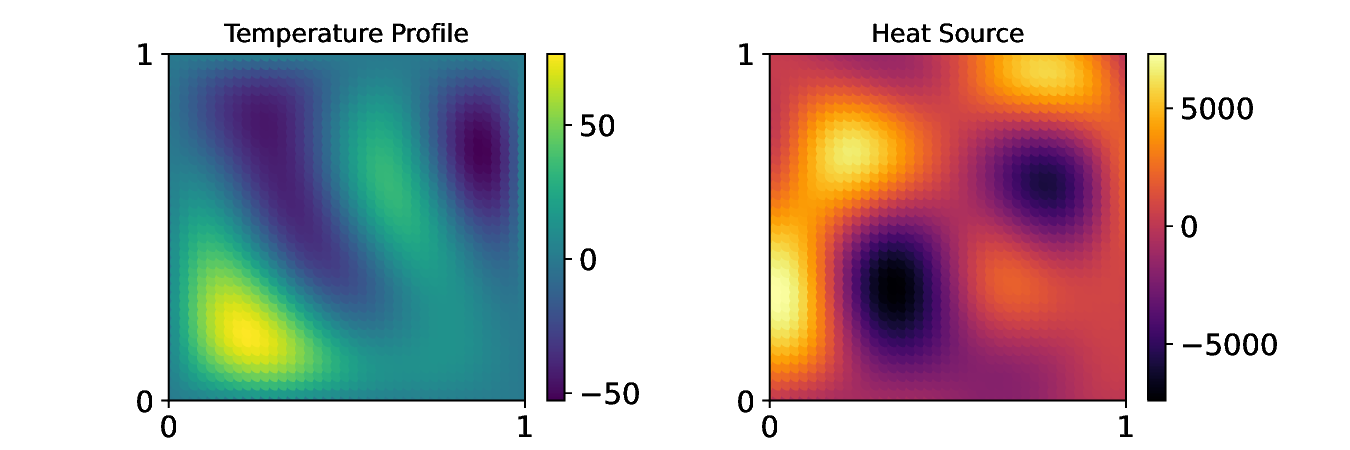}
        \caption{\label{6_2tb}}
    \end{subfigure}
    \begin{subfigure}[b]{0.33\textwidth}
        \centering
        \includegraphics[trim= {2cm 0 12cm .75cm},clip, width=1\textwidth]{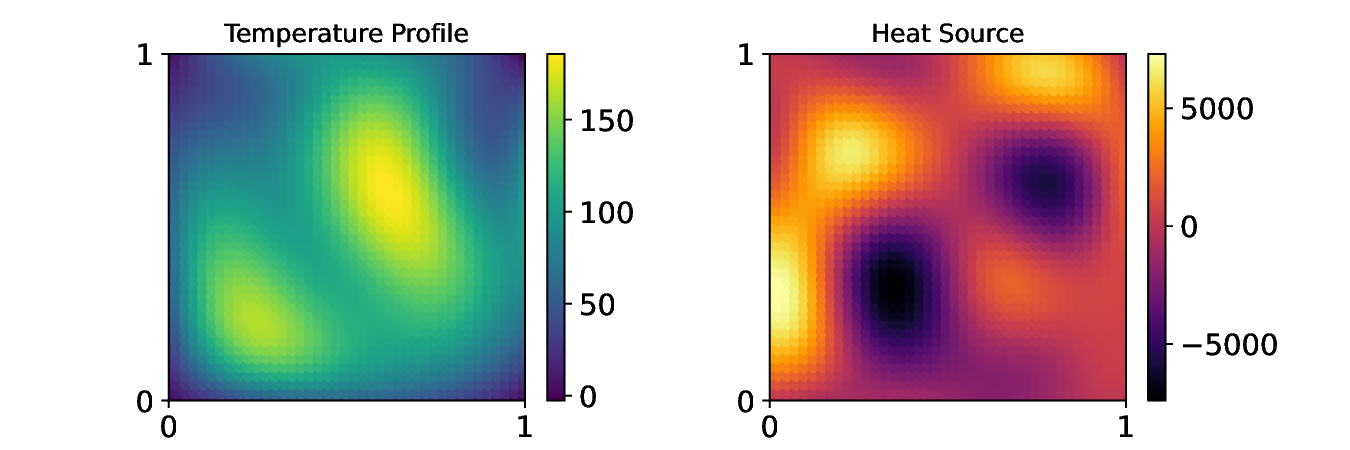}
        \caption{\label{6_2tc}}
    \end{subfigure}
    \caption{Generation of the temperature data for heterogeneous boundary conditions [\(T_{x=0,y}=200y(1-y), T_{x=1,y}=400y(1-y), T_{x,y=0}=200x(1-x), T_{y=1}=400x (1-x)\) :  (\subref{6_2ta}) profile satisfying non-homogeneous boundary condition, (\subref{6_2tb}) profile satisfying homogeneous boundary condition,  (\subref{6_2tc}) final profile obtained by addition of profile (a) and (b). }
    \label{6_2t}
\end{figure}

The temperature values at the grid points can be generated from the above posterior distribution data. Note that the value of the \(\sigma_{\epsilon}^2\) should be taken to be small, in order to satisfy the boundary conditions with high accuracy. In our numerical examples, we have taken the typical value of the order of $10^{-3}$. Thus, the data generated from this posterior distribution satisfy at the boundary points equals to the boundary conditions while is random in nature at the interior points. A sample representation for the same can be seen in \autoref{5_0} where GPR is used to generate random fields which satisfy the homogeneous boundary conditions.
\\We observed that for homogeneous boundary conditions i.e. \(T=0\) at all boundaries, in general, the data generated from a single length parameter value is sufficient. However, for data generation for the heterogeneous boundary case, multiple length scales are required. In this manuscript, we propose that, we can generate a  heterogeneous profile which satisfies the boundary condition with larger length scale. Further, the data with homogeneous boundary condition and the smaller length scale can be generated. All the homogeneous profiles are added with the heterogeneous profile to obtain multiple temperature profiles. From this methodology, we can obtain the variance in the temperature field from the homogeneous profiles with smaller length scale, while the boundary conditions are satisfied from the heterogeneous profile with large length scale. For example, a typical random temperature profile for a square domain for the heterogeneous boundary condition is shown in the~\autoref{6_2t}.  In this figure, the  heterogeneous profile is generated with  GPR\((l=4,{\sigma}=20)\), while the homogeneous profile is generated with \((l=0.3, \sigma=40)\). As can be seen from the figure that the final profile has the variation of the smaller length scale while satisfying the heterogeneous boundary conditions.
%addition of the heterogeneous profile with all of the 

\subsection{Calculation of the random heat source}
To calculate the heat source, above generated random temperature field is substituted in to the partial differential equation. The technique of central difference is employed to calculate the partial derivatives. 
\begin{align}
     f_{,x}(x_i,y_i) &= \frac{f(x_{i+1},y_i) - f(x_{i-1},y_i)}{2h}\\
     f_{,y}(x_i,y_i) &= \frac{f(x_i,y_{i+1}) - f(x_i,y_{i-1})}{2h}
     \label{fdm}
\end{align}
where, $(x_{i+1},y_{i+1})$ and $(x_{i-1},y_{i-1})$ are neighboring points of point $(x_i,y_i)$ and $2h$ refers to the distance between the points $i-1$ and $i+1$. 
%Since the quantities on the right hand side are obtained from the Gaussian random field, the computed derivatives being a linear combination of the right hand quantities are also going to be a Gaussian random field in nature.  
Thus after calculating these partial derivatives, the heat flux at the grid points can be simply calculated as below:
\begin{equation}
    q(x,y) = - (kT_{,x})_{,x} - (kT_{,y})_{,y}
    \label{heats}
\end{equation}
  Thus from the above proposed startegy, the pair of temperature- heat flux are obtained. Thess values of the temperature field and the corresponding obtained heat source values can be used for training the operator network.   Note in numerical simulation, we omit some of the grid points near the domain boundary. This is done due to noisy nature of the derivatives near the boundaries.

%Continuing with the details of the methodology, the subsequent sections present the details about the generation of the random temperature field.   

\subsection{Note on computational complexity}
The most computationally intensive operation in the proposed framework is the inversion of the \((K + \sigma_{\epsilon}^2I)\) matrix in~\autoref{eq_gpr} with \(n_o\times n_o\) dimension. While in case of the technique such as FEM the size of the matrix require inversion is \(n\times n\). Where, \(n\) is the total number of points in the grid. Since, the overall number of the points on the boundary \(n_0\) is going to be significantly less compared to the overall points \(n\) in the domain, GPR is expected to take significantly less computationally efforts compared to FEM. 
In summary, in the proposed framework, we have obtained the discretized temperature fields satisfying the specified boundary conditions along with the discretized values of the heat source in a computationally efficient manner.

\section{Numerical Examples}\label{Results}
In the previous sections, a general strategy that reduces the cost of generating training data for DeepONet for any given domain and boundary condition for a PDE was proposed. In this section, we present various numerical examples to demonstrate the effectiveness of the DeepONet based surrogate model trained using our proposed approach. We will be using 2-D Poisson steady-state heat conduction equation (\autoref{e_2_2}) for validating our approach by comparing the prediction of our model with analytical results obtained using FEM analysis. In all the examples, we  have considered only the Dirichlet Boundary Conditions. %For simplicity, we have taken the $k$ to be either equal to $1$ or $1+0.01T$.  

The dataset for training the DeepONet have been prepared with the proposed methodology. The dataset was split into an 80-20 ratio for the purpose of training and testing. The DeepONet was trained using mean squared error as the loss function.  In all numerical examples we report the results with R2 score and Normalised L2(\autoref{e_l2}) metric. 
\begin{equation}
    \text{Normalized L2} = \frac{1}{N_p} \frac{\|T_{\text{predicted}} - T_{\text{actual}}\|_2}{\|T_{\text{actual}}\|_2}.
\label{e_l2}
\end{equation}
GlorotNormal initializer has been used to initialize weights and relu activation function is used in the hidden layers. For further validation of our DeepONet model, the following three heat source are considered:
\begin{equation}
\begin{aligned}
    q_1(x,y) &= a,\\
    q_2(x,y) &= b\sin(x)\sin(y),\\
    q_3(x,y) &= c\left( \sin(5x)\sin(3y) + \cos(5y)\cos(2x) \right).
\end{aligned}
\label{testf}
\end{equation}
Here, \(a,b, \text{c}\) are constants, whose values are mentioned in the specific numerical examples.
\subsection{Square domain with constant thermal conductivity}
In this numerical example, we have considered a square domain with a side length of one. The domain is discretized with $41\times41$ uniformly distributed points within the domain and thermal conductivity $k(x,y) = 1$. For this example, we have developed the surrogate model using DeepONet for homogeneous as well as heterogeneous boundary conditions. For training and testing purpose a total dataset of \(200\) temperature profiles is considered.

\subsubsection{Homogeneous Boundary Condition}
\begin{figure}[h]
    \centering
    \includegraphics[trim= {1cm 0 1cm 0},clip, width=0.65\textwidth]{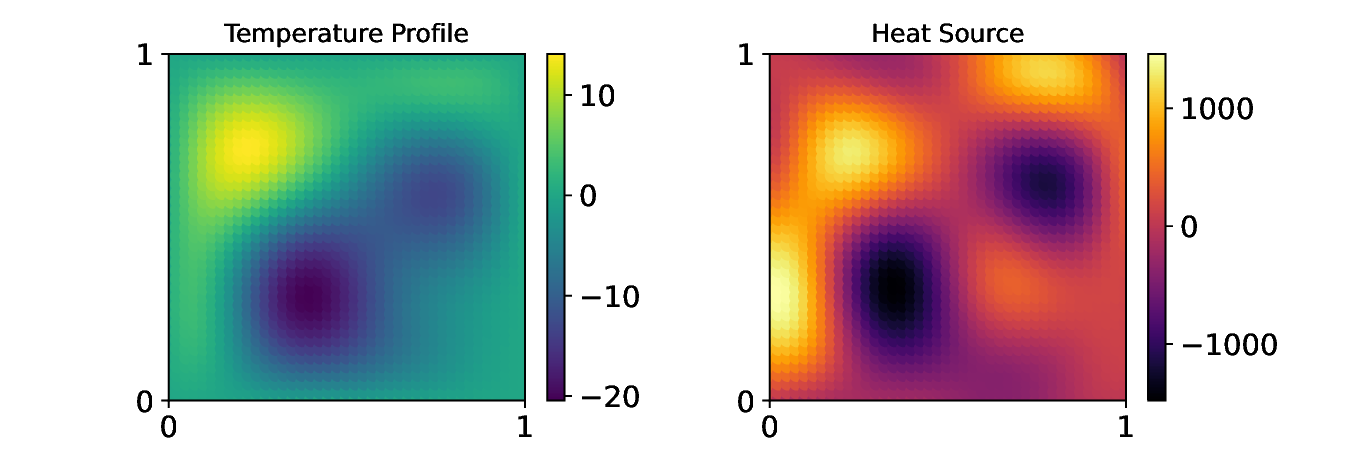}
    \caption{Square domain with homogeneous boundary condition: A typical temperature and heat source profile obtained using the proposed framework}
    \label{6_0}
\end{figure}

\begin{figure}[ht!]
\centering
    \begin{subfigure}[b]{1\textwidth}
        \centering
        \includegraphics[trim= {5.5cm 0.41cm 5cm 0.45cm},clip, width=1\textwidth]{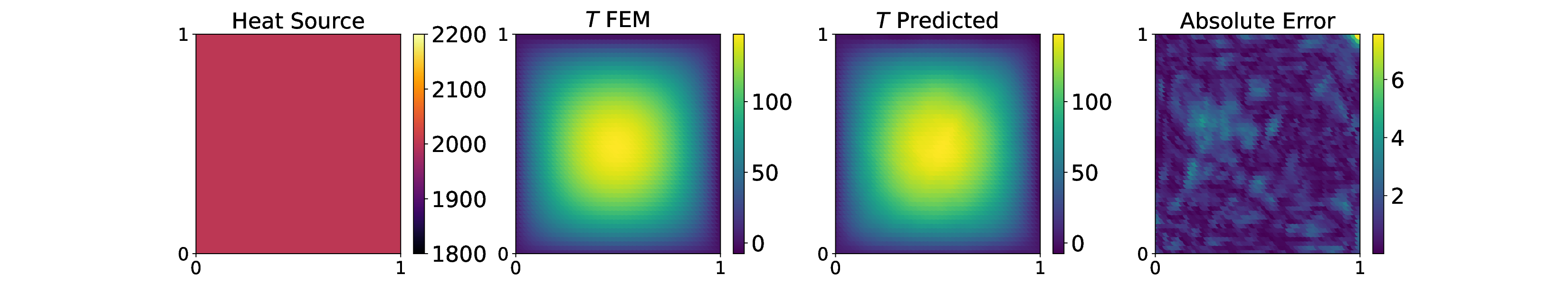}
        \caption{\label{6_1a}}
    \end{subfigure}
    \begin{subfigure}[b]{1\textwidth}
        \centering
        \includegraphics[trim= {5.5cm 0.41cm 5cm 0.45cm},clip, width=1\textwidth]{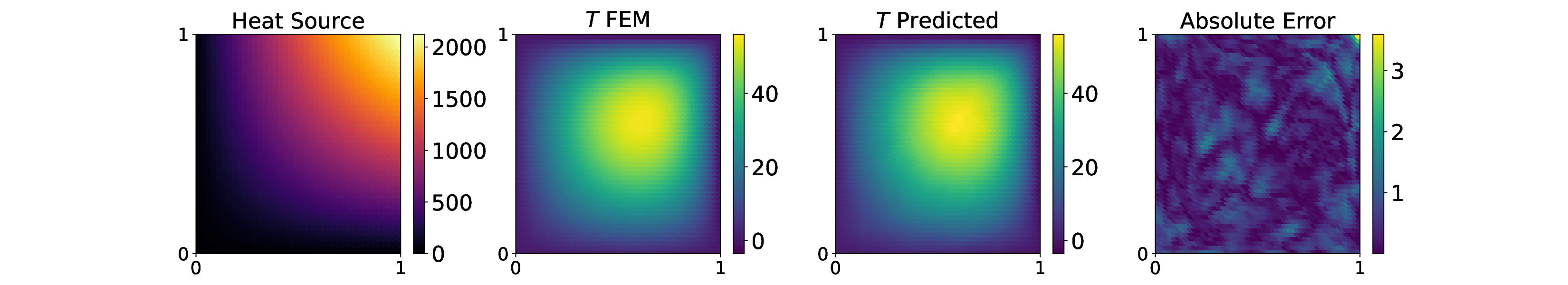}
        \caption{\label{6_1b}}
    \end{subfigure}
    \begin{subfigure}[b]{1\textwidth}
        \centering
        \includegraphics[trim= {5.5cm 0.41cm 5cm 0.45cm},clip, width=1\textwidth]{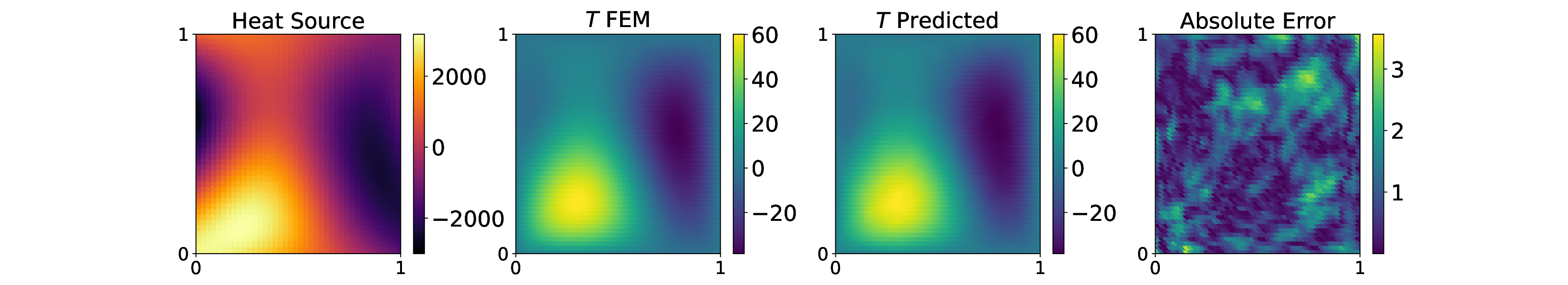}
        \caption{\label{6_1c}}
    \end{subfigure}
\caption{Homogeneous boundary condition with constant thermal conductivity: DeepONet predictions for various heat source test functions compared with the FEM data: (\subref{6_1a}) $q(x,y)$=$2000$ , (\subref{6_1b})  $q(x,y)$=$3000\sin(x)\sin(y)$, and (\subref{6_1c})  $q(x,y)$=$3000\left( \sin(5x)\sin(3y) + \cos(5y)\cos(2x) \right)$ . The R2 score of these three cases are 0.99934, 0.9989, \text{and} 0.997918 respectively.} 
\label{6_1}
\end{figure}

For homogeneous boundary condition, we have set temperature \(T=0\), at all boundaries. For generating the random temperature profiles satisfying the homogeneous boundary conditions, we have used GPR with length scale parameter as $l=0.3$ and $\sigma=50$. From these temperature profiles the heat source is calculated using the central difference scheme as described in section.  For reference, one typical random temperature and the corresponding heat flux profile is shown in~\autoref{6_0}.\\
\\The DeepONet architecture features zero hidden layers in branch net and four hidden layers (150 neurons in each layer) in trunk net. The output layer of branch net and trunk net consists of the 200 neurons. The DeepONet model is trained using 200 epochs with a batch size of 160. We have used the decaying learning rate where the specific parameters are: (start = 0.0001, decay rate = 0.96 and decay steps = 1000). The DeepONet model obtained is having a R2 score of \textbf{0.99986} on the testing data set. Further, we test the trained model for the three test function given in \autoref{testf}. The value of the constants a, b and c are taken as 2000, 3000 and 3000 respectively. The temperature profile predictions for these three test cases using our trained model are compared with the FEM results in \autoref{6_1}. As can be seen from the figure, the DeepONet predictions shows a good match with FEM results both qualitatively as well as quantitatively. We have also calculated the L2 error for these cases. The L2 error for these cases are \(5.4\times10^{-6}\), \(7.65\times10^{-6}\) , \(1.739\times10^{-5}\)respectively. As can be seen that predictions from the DeepONet model have quite low L2 error, thus further validating the accuracy of the obtained model. 

\subsubsection{Heterogeneous Boundary Condition}
\begin{figure}[ht!]
    \centering
    \begin{subfigure}[b]{0.49\textwidth}
        \centering
        \includegraphics[trim= {1cm 0 1cm 0},clip, width=1\textwidth]{Figures/results/square/heterogeneous/nonhomtrainfunc.eps}
        \caption{\label{6_2tta}}
    \end{subfigure}
    \begin{subfigure}[b]{0.49\textwidth}
        \centering
        \includegraphics[trim= {1cm 0 1cm 0},clip, width=1\textwidth]{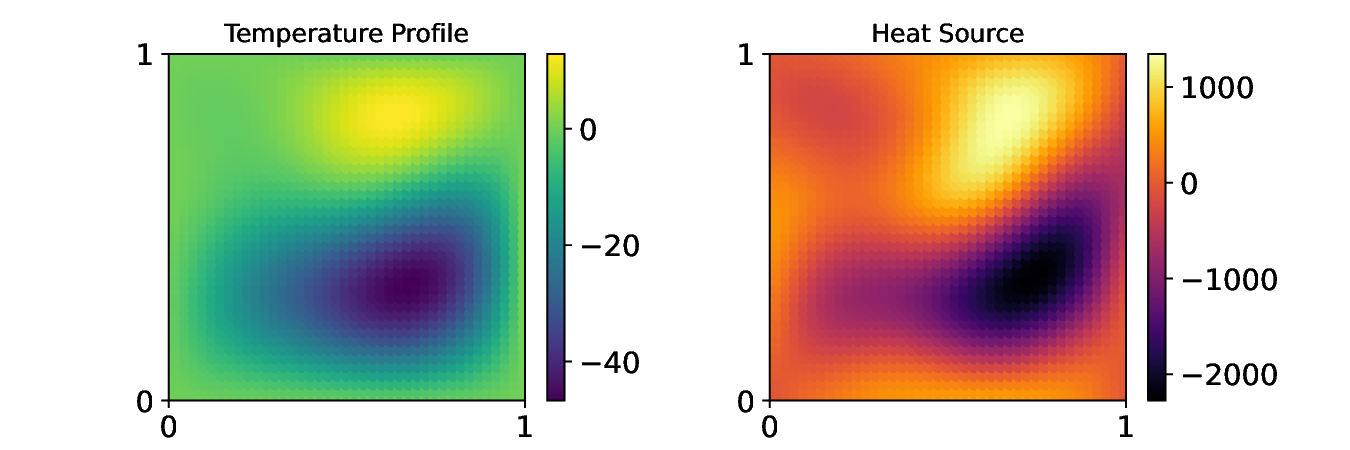}
        \caption{\label{6_2ttb}}
    \end{subfigure}
    \begin{subfigure}[b]{1\textwidth}
        \centering
        \includegraphics[trim= {1cm 0 1cm 0},clip, width=0.5\textwidth]{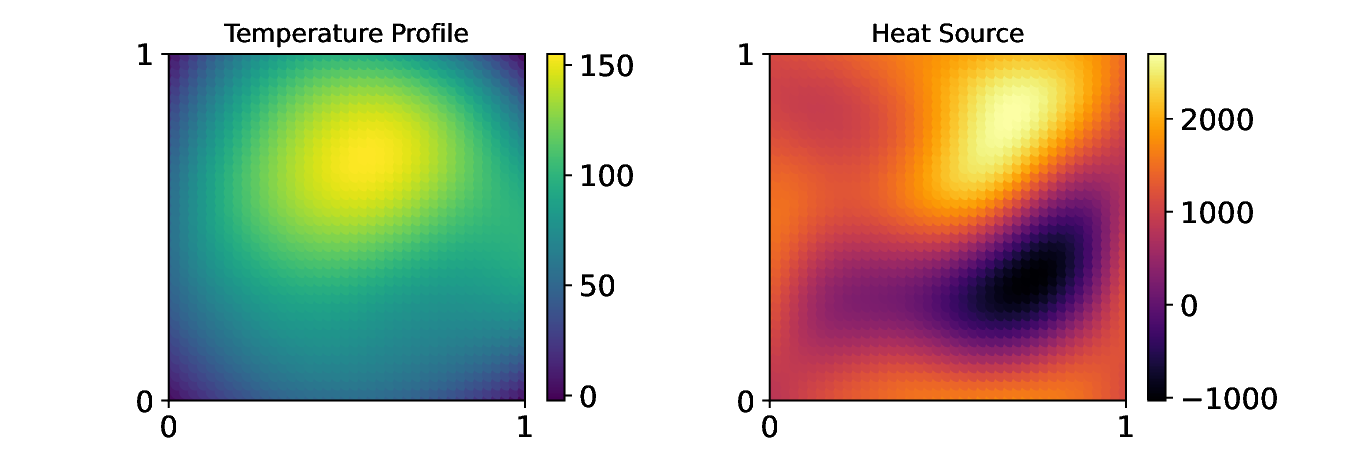}
        \caption{\label{6_2ttc}}
    \end{subfigure}
    \caption{Generation of the temperature data for heterogeneous boundary conditions:  (\subref{6_2tta}) profile satisfying non-homogeneous boundary condition, (\subref{6_2ttb}) profile satisfying homogeneous boundary condition,  (\subref{6_2ttc}) final profile obtained by addition of profile (a) and (b). The heat source obtained using the finite difference from these profiles are also shown. }
    \label{6_2tt}
\end{figure}

\begin{figure}[ht!]
\centering
    \begin{subfigure}[b]{1\textwidth}
        \centering
        \includegraphics[trim= {5.5cm 0.41cm 5cm 0.45cm},clip, width=1\textwidth]{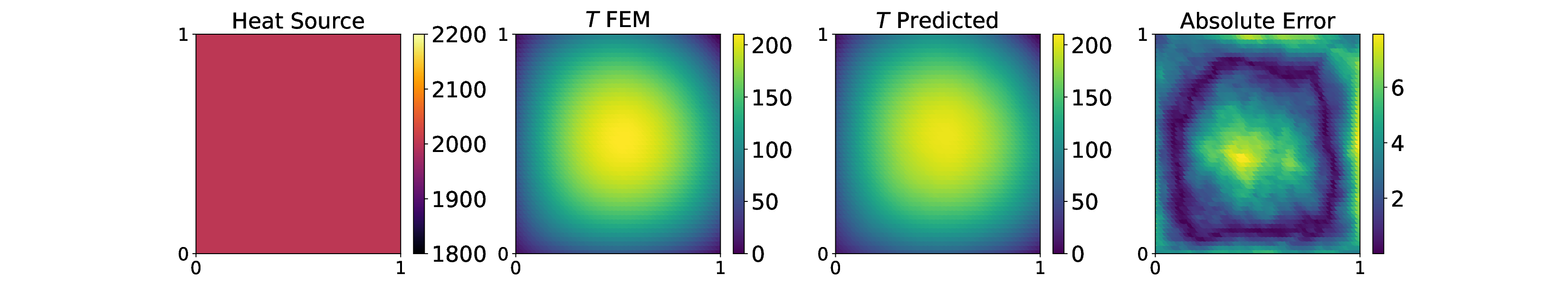}
        \caption{\label{6_2a}}
    \end{subfigure}
    \begin{subfigure}[b]{1\textwidth}
        \centering
        \includegraphics[trim= {5.5cm 0.41cm 5cm 0.45cm},clip, width=1\textwidth]{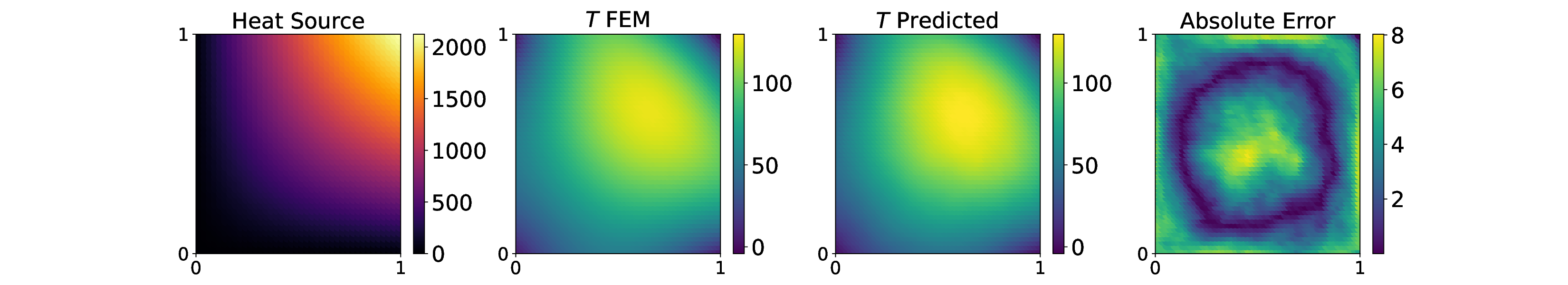}
        \caption{\label{6_2b}}
    \end{subfigure}
    \begin{subfigure}[b]{1\textwidth}
        \centering
        \includegraphics[trim= {5.5cm 0.41cm 5cm 0.45cm},clip, width=1\textwidth]{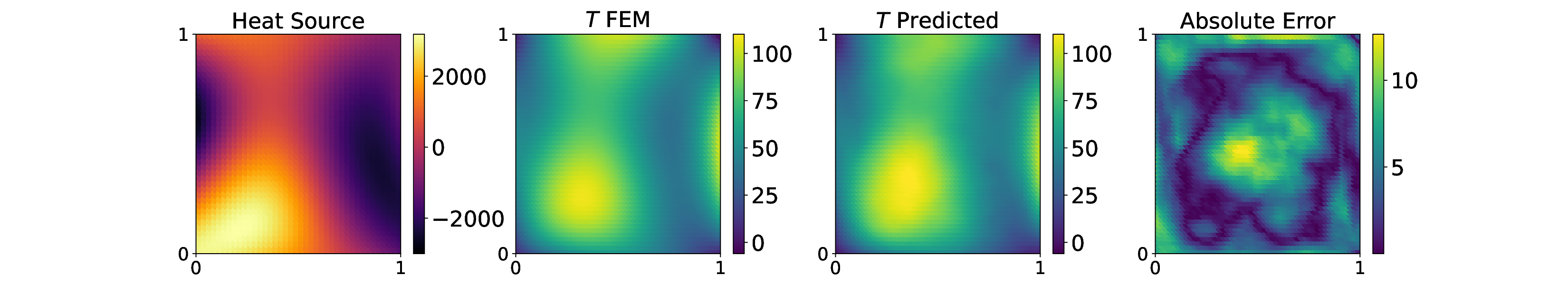}
        \caption{\label{6_2c}}
    \end{subfigure}
\caption{Heterogeneous boundary condition with constant thermal conductivity: DeepONet predictions for various heat source test functions compared with the FEM data: (\subref{6_2a}) $q(x,y)$=$2000$ , (\subref{6_2b})  $q(x,y)$=$3000\sin(x)\sin(y)$, and (\subref{6_2c})  $q(x,y)$=$3000\left( \sin(5x)\sin(3y) + \cos(5y)\cos(2x) \right)$ . The R2 score of these three cases are 0.9947, 0.9843, \text{and} 0.9516 respectively.}
\label{6_2}
\end{figure}

The boundary condition selected for heterogeneous case are chosen as: \(T_{x=0,y}=200y(1-y) \text{ [bottom edge]}, T_{x=1,y}=400y(1-y) \text{ [top edge]}, T_{x,y=0}=200x(1-x) \text{ [left edge]}, T_{y=1}=400x (1-x)\) \text{ [right edge]}.
For heterogeneous boundary condition, we have generated the data with two length scales. First, we have generated a single heterogeneous profile which satisfied the boundary condition with GPR\((l=4,{\sigma}=20)\). Subsequently,  200 homogeneous temperature profiles have been generated with \(T=0\) at all boundaries with GPR \((l=0.3, \sigma=40)\). The addition of  heterogeneous profile to each of the homogeneous temperature profiles has been performed to obtain the final profiles. A typical sample training data in \autoref{6_2tt} shows that the temperature field has the variation similar to the homogeneous profiles~\autoref{6_0}, while satisfying the heterogeneous boundary conditions also.
\\The DeepONet architecture features zero hidden layers in branch net and four hidden layers (150 neurons in each layer) in trunk net with 200 neurons in the common layer. The model was trained for 200 epochs with a batch size of 160 and with a decaying learning rate (start = 0.0001, decay rate = 0.96 and decay steps = 1000). The trained model shows a R2 score of \textbf{0.99986} on the test set. For  additional testing of our model, we use the heat source functions with \(a=2000,\text{ }b=c=3000\) given in ~\autoref{testf}. The temperature field obtained  for these set of functions are shown in \autoref{6_2}. The L2 error for these cases are \(1.0039\times10^{-5}\), \(1.80788\times10^{-5}\) , \(3.0947\times10^{-5}\)respectively.

As can be seen from the figure, in the case of the heterogeneous boundary conditions also the DeepONet predictions shows a good match with FEM results.

\subsection{Square domain with varying thermal conductivity}
\begin{figure}[ht!]
    \centering
    \includegraphics[trim= {1cm 0 1cm 0},clip, width=0.65\textwidth]{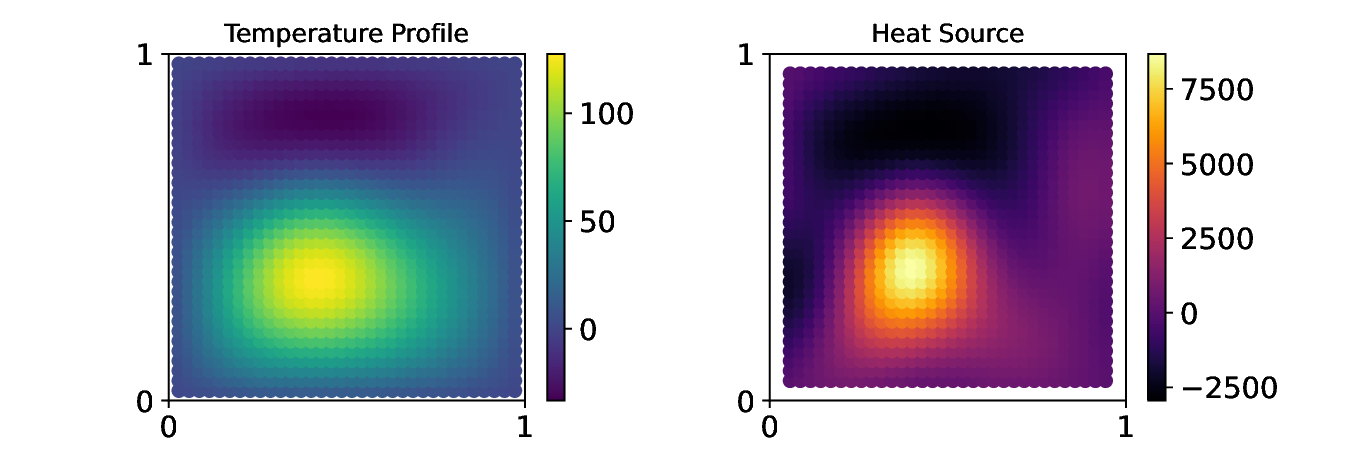}
    \caption{A typical temperature and heat source profile obtained for square domain with varying thermal conductivity using the proposed methodology.}
    \label{6nl}
\end{figure}

\begin{figure}[ht!]
\centering
    \begin{subfigure}[b]{1\textwidth}
        \centering
        \includegraphics[trim= {5.5cm 0.3cm 5cm 0.43cm},clip, width=1\textwidth]{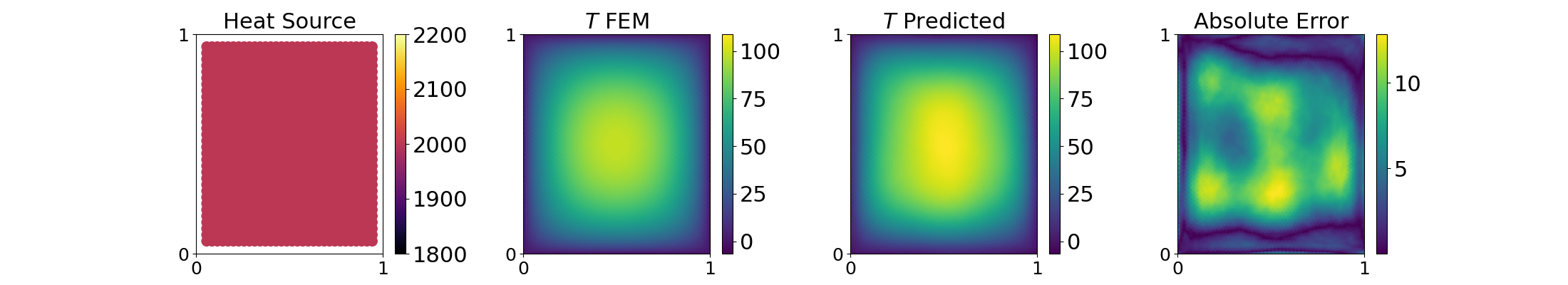}
        \caption{\label{6_1nla}}
    \end{subfigure}
    \begin{subfigure}[b]{1\textwidth}
        \centering
        \includegraphics[trim= {5.5cm 0.3cm 5cm 0.43cm},clip, width=1\textwidth]{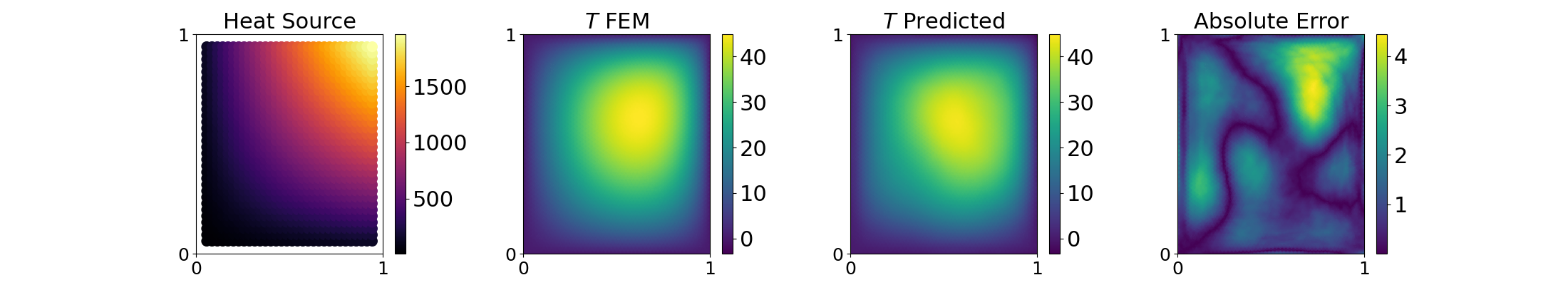}
        \caption{\label{6_1nlb}}
    \end{subfigure}
    \begin{subfigure}[b]{1\textwidth}
        \centering
        \includegraphics[trim= {5.5cm 0.3cm 5cm 0.43cm},clip, width=1\textwidth]{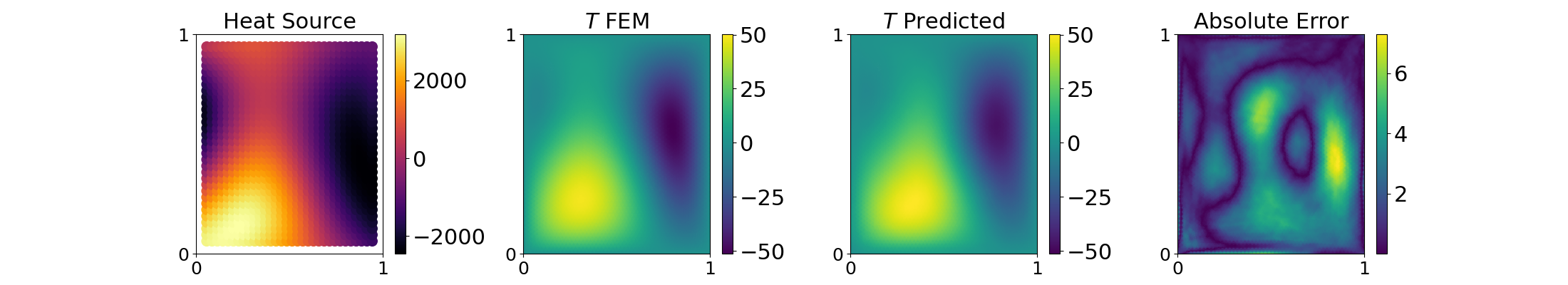}
        \caption{\label{6_1nlc}}
    \end{subfigure}
\caption{Homogeneous boundary condition with variable thermal conductivity: DeepONet predictions for various heat source test functions compared with the FEM data: (\subref{6_1nla}) $q(x,y)$=$2000$ , (\subref{6_1nlb})  $q(x,y)$=$3000\sin(x)\sin(y)$, and (\subref{6_1nlc})  $q(x,y)$=$3000\left( \sin(5x)\sin(3y) + \cos(5y)\cos(2x) \right)$ . The R2 score of these three cases are 0.957557, 0.98481, \text{and} 0.9863088 respectively.}
\label{6_1nl}
\end{figure}

In this numerical example, we have considered a square domain with  thermal conductivity $k(x,y) = 1+0.01T(x,y)$. The domain is discretized into $1024$ uniformly distributed points within the domain. We set the boundary condition as $T=0$ along all boundaries and generate a data set of \(1000\) heat source and temperature profiles. To generate random temperature profiles that adhere to the boundary conditions, we employ a mixture of profiles using Gaussian Process Regression (GPR). 
Specifically, 800 profiles are generated using GPR with a length scale parameter as $l=0.3$ and $\sigma=50$, while 200 profiles are created with GPR using a length scale parameter as $l=0.3$ and $\sigma=80$. The model is first trained on the profiles of low sigma value. The obtained model is taken as the initial guess and then further trained on the data with $\sigma=80$.  For reference, one typical random temperature and the corresponding heat flux profile obtained with $l=0.3$ and $\sigma=80$ is shown in~\autoref{6nl}.\\
\\The DeepONet architecture features two hidden layers in branch net and four hidden layers in trunk net with 300 neurons in each hidden layer. The output layer of branch net and trunk net consists of the 350 neurons. The DeepONet model is trained using 200 epochs where we used 0.001 as the learning rate for first 10 epochs, 0.0001 learning rate for next 100 epochs and then 0.00005 learning rate for remaining epochs. The DeepONet model obtained is having  a R2 score of \textbf{0.9943279} on the testing data set. For  additional testing of our DeepONet model, we take heat source function mentioned in \autoref{testf} with $a=2000$ and $b=c=3000$. The comparison of temperature predictions for these test cases using our trained model with the FEM results is shown in \autoref{6_1nl}. The L2 error for these cases are \(3.92\times10^{-6}\), \(2.31538\times10^{-6}\) , \(4.00925\times10^{-6}\) respectively. 

\subsection{Triangular domain}
\begin{figure}[ht!]
    \centering
    \includegraphics[trim= {2.25cm 0 1.25cm 0},clip, width=0.65\textwidth]{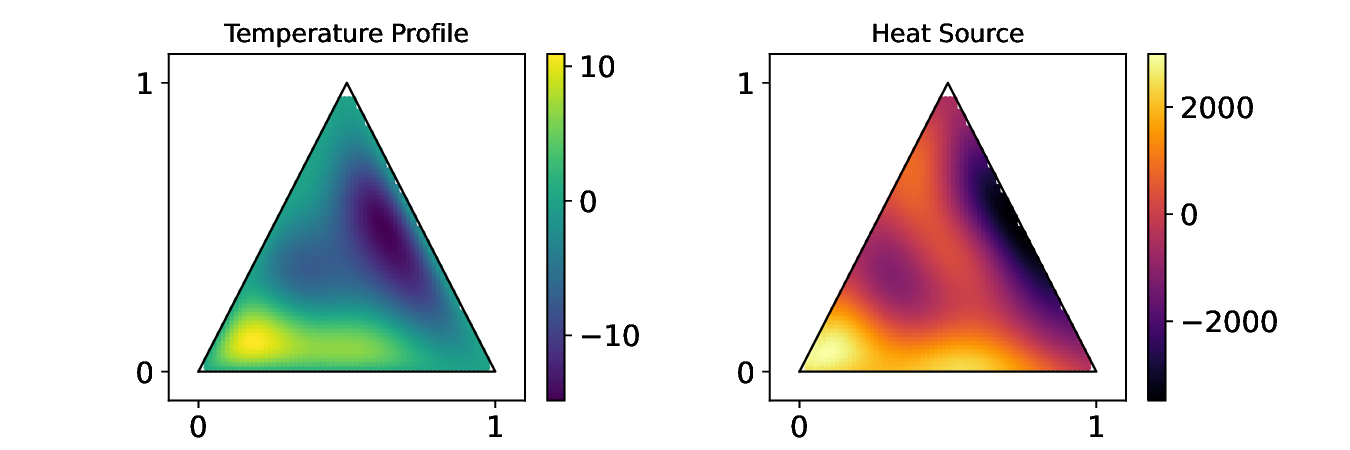}
    \caption{A typical temperature and heat source profile obtained for triangular domain using the proposed framework.}
    \label{6_3}
\end{figure}

\begin{figure}[ht!]
\centering
    \begin{subfigure}[b]{1\textwidth}
        \centering
        \includegraphics[trim= {5.5cm 0.3cm 5cm 0.43cm},clip, width=1\textwidth]{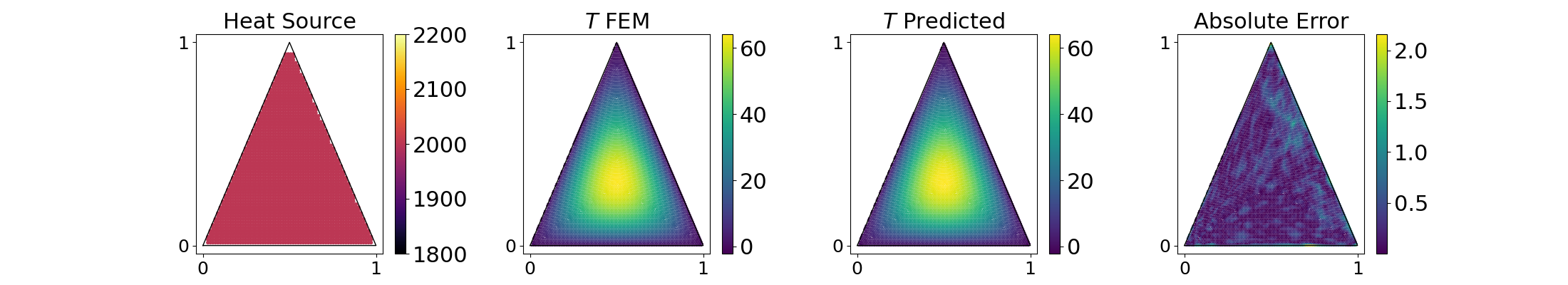}
        \caption{\label{6_4a}}
    \end{subfigure}
    \begin{subfigure}[b]{1\textwidth}
        \centering
        \includegraphics[trim= {5.5cm 0.3cm 5cm 0.43cm},clip, width=1\textwidth]{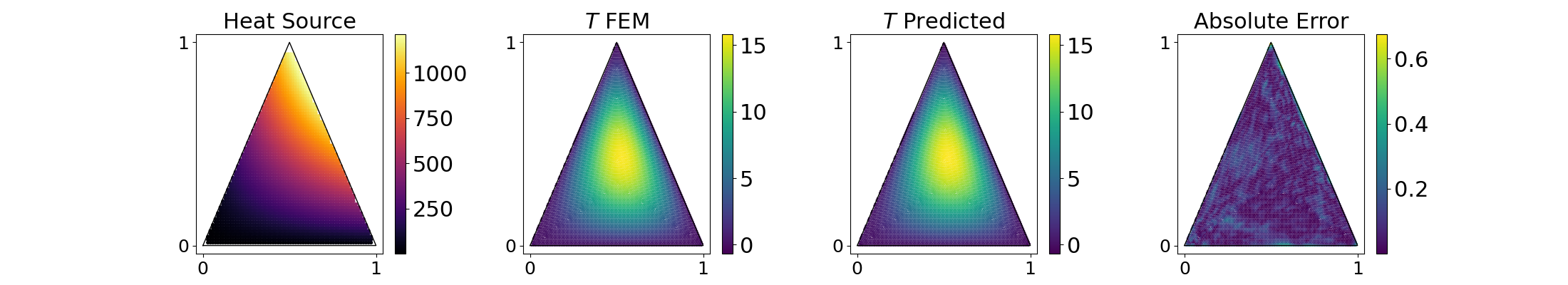}
        \caption{\label{6_4b}}
    \end{subfigure}
    \begin{subfigure}[b]{1\textwidth}
        \centering
        \includegraphics[trim= {5.5cm 0.3cm 5cm 0.43cm},clip, width=1\textwidth]{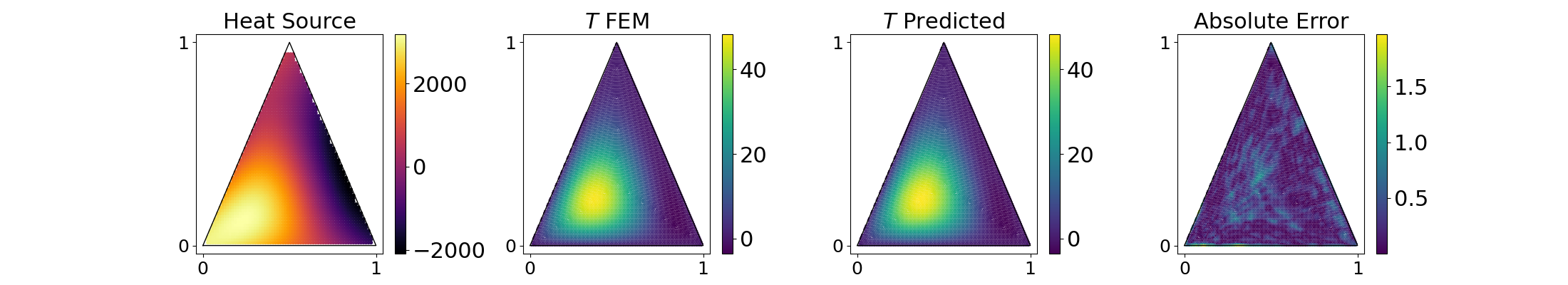}
        \caption{\label{6_4c}}
    \end{subfigure}
\caption{Homogeneous boundary condition with constant thermal conductivity for triangular domain: DeepONet predictions for various heat source test functions compared with the FEM data: (\subref{6_4a}) $q(x,y)$=$2000$ , (\subref{6_4b})  $q(x,y)$=$3000\sin(x)\sin(y)$, and (\subref{6_4c})  $q(x,y)$=$3000\left( \sin(5x)\sin(3y) + \cos(5y)\cos(2x) \right)$ . The R2 score of these three cases are 0.9997, 0.999628, \text{and} 0.9995758 respectively.}
\label{6_4}
\end{figure}

In this example, we consider a triangular domain having vertices at (0,0),(1,0) and (0.5,1). The domain is discretized into 2299 uniformly distributed points within the domain and thermal conductivity $k(x,y) = 1$. We have set $T=0$ at all boundaries as the boundary condition to generate a total of 200 temperature profiles. The temperature data was generated using GPR with $l=0.3$ and $\sigma=150$. The training data is generated using the methodology proposed. The DeepONet architecture features zero hidden layers in branch net and four hidden layers (200 neurons in each layer) in trunk net with 150 neurons in the common layer. A sample training data can be seen in \autoref{6_3}.
\\The model is trained for 100 epochs with a batch size of 1024 and with a decaying learning rate (start = 0.0001, decay rate = 0.96 and decay steps = 1000). The DeepONet model has a R2 score of \textbf{0.99995} on the testing dataset. For additional testing of our DeepONet model, we take heat source function mentioned in \autoref{testf} with $a=2000$ and $b=c=3000$. The comparison of temperature predictions for these test cases using our trained model with the FEM results is shown in \autoref{6_4}. The L2 error for these cases are \(1.505\times10^{-6}\), \(1.811968\times10^{-6}\), \(2.37605\times10^{-6}\) respectively.
\subsection{Annular domain}
\begin{figure}[ht!]
    \centering
    \includegraphics[trim= {2.25cm 0 1.25cm 0},clip, width=0.65\textwidth]{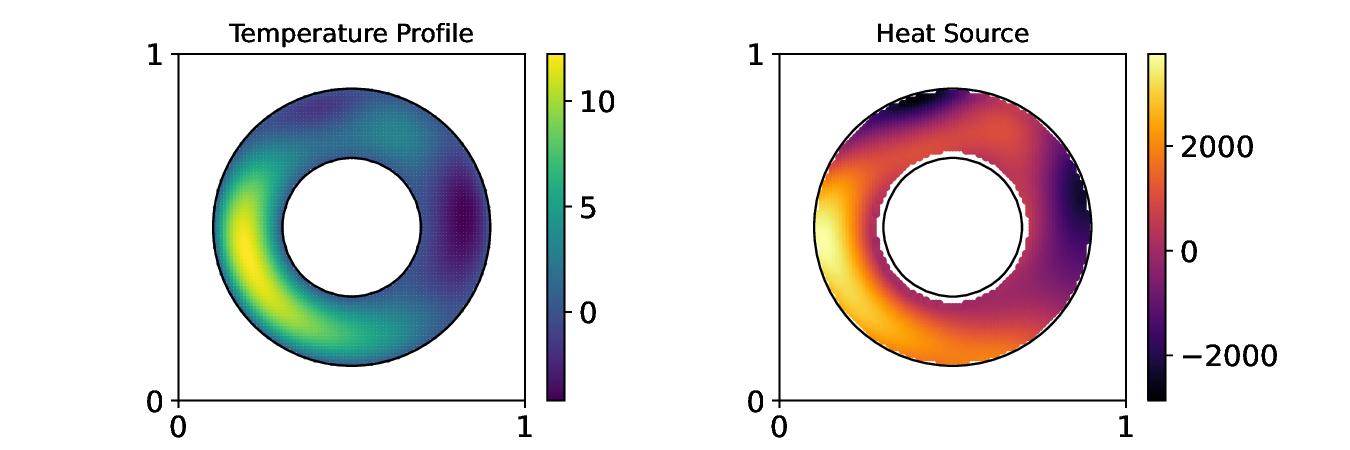}
    \caption{A typical temperature and heat source profile obtained for annular profile using the proposed framework.}
    \label{6_5}
\end{figure}

For this example, we consider an annular domain with centre at (0.5,0.5) and inner and outer radius of 0.2 and 0.4 units respectively. The domain is discretized into 3404 uniformly spaced points within the domain and thermal conductivity $k(x,y) = 1$. We have set $T=0$ at all boundaries as the boundary condition to generate a total of 200 temperature profiles. The temperature data was generated using GPR with $l=0.2$ and $\sigma=60$. The training data is generated using the methodology proposed.The DeepONet architecture features zero hidden layers in branch net and four hidden layers (150 neurons in each layer) in trunk net with 200 neurons in the common layer. A sample training data can be seen in \autoref{6_5}.
\begin{figure}[ht!]
\centering
    \begin{subfigure}[b]{1\textwidth}
        \centering
        \includegraphics[trim= {5.5cm 0.8cm 5cm 1cm},clip, width=1\textwidth]{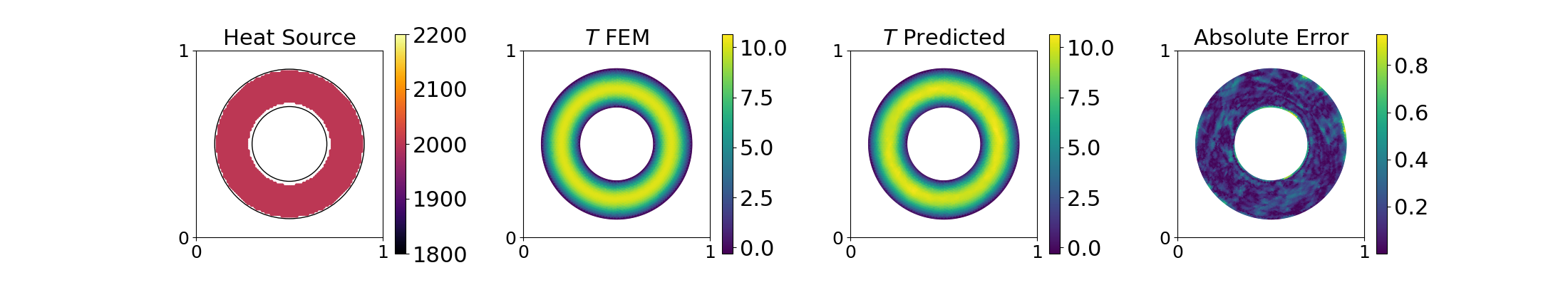}
        \caption{\label{6_6a}}
    \end{subfigure}
    \begin{subfigure}[b]{1\textwidth}
        \centering
        \includegraphics[trim= {5.5cm 1cm 5cm 1cm},clip, width=1\textwidth]{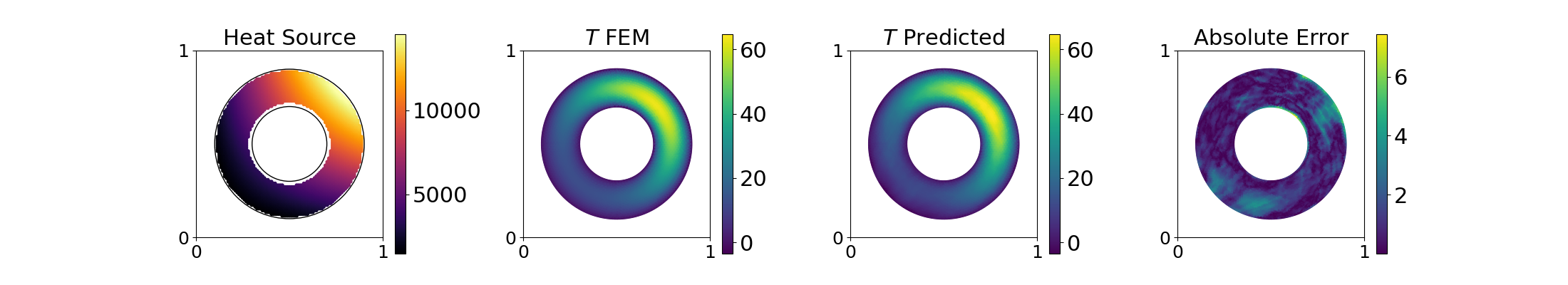}
        \caption{\label{6_6b}}
    \end{subfigure}
    \begin{subfigure}[b]{1\textwidth}
        \centering
        \includegraphics[trim= {5.5cm 1cm 5cm 1cm},clip, width=1\textwidth]{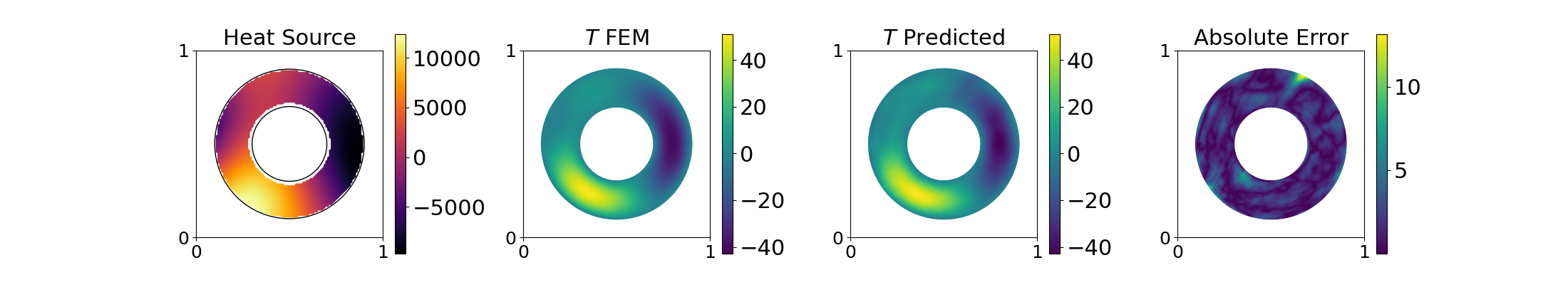}
        \caption{\label{6_6c}}
    \end{subfigure}
\caption{Homogeneous boundary condition with constant thermal conductivity for annular domain: DeepONet predictions for various heat source test functions compared with the FEM data: (\subref{6_6a}) $q(x,y)$=$2000$ , (\subref{6_6b})  $q(x,y)$=$12000\sin(x)\sin(y)$, and (\subref{6_6c})  $q(x,y)$=$30000\left( \sin(5x)\sin(3y) + \cos(5y)\cos(2x) \right)$ . The R2 score of these three cases are 0.99644, 0.99365, \text{and} 0.989743 respectively.}
\label{6_6}
\end{figure}
\\The model is trained for 200 epochs with a batch size of 1024 and with a decaying learning rate (start = 0.0001, decay rate = 0.96 and decay steps = 1000). Once trained, the model has a R2 score of \textbf{0.99937} on the test set. For  additional testing of our DeepONet model, we take heat source function mentioned in \autoref{testf} with $a=2000$, $b=12000$ and $c=30000$. The comparison of temperature predictions for these test cases using our trained model with the FEM results is shown in \autoref{6_6}. The L2 error for these cases are \(1.111814\times10^{-6}\), \(2.34158\times10^{-6}\) , \(4.8036\times10^{-6}\)respectively.

 As can be seen from the above results, the DeepONet predictions shows a good match with FEM results both qualitatively as well as quantitatively which depicts that DeepONet based surrogate model can be trained for any arbitrary shaped domain without using computationally expensive simulations from FEM.

\section{Conclusions}\label{conclusion}
In this work, we have proposed a novel data generation framework data-driven approach for approximating the solution of governing partial different equation of physical systems using DeepONet. In the proposed framework, we have used GPR in conjunction with finite difference technique towards generation of the training data. The proposed approach have been shown to be  applicable for any arbitrary shaped domain and boundary conditions.   Overall, our work represents a significant step towards advancing data-driven approaches for surrogate modeling, paving the way for more efficient modeling of complex physical systems.

%----------------------------------------------------------------
%				 		add references 
%----------------------------------------------------------------

\renewcommand{\bibname}{References}		% changes default name Bibliography to References

\bibliographystyle{elsarticle-num}				% define reference style
\bibliography{references}	% call the .bib file for references
% ---------------------------------------------

\end{document}